\documentclass[letterpaper, 10 pt, journal, twoside]{ieeetran}  

\IEEEoverridecommandlockouts                              

\title{I've Changed My Mind: Robots Adapting to Changing Human Goals during Collaboration
}

\author{Debasmita Ghose$^{*}$, Oz Gitelson$^{*}$, Ryan Jin, Grace Abawe, Marynel V\'azquez, and Brian Scassellati%
\thanks{Manuscript received: June, 24, 2024; Revised October, 10, 2025; Accepted November, 19, 2025.}
\thanks{This paper was recommended for publication by Editor Angelika Peer upon evaluation of the Associate Editor and Reviewers' comments.
This work was supported by the National Science Foundation (NSF) awards IIS-2106690 and IIS-1955653, and Office of Naval Research (ONR) award N00014-24-1-2124.) Any opinions, findings, conclusions, or recommendations expressed in this material are those of the authors and do not
necessarily reflect the views of the NSF or ONR.} 
\thanks{$^{*}$Authors Contributed Equally. The authors are with the Department of Computer Science, Yale University, New Haven, CT, USA
        {\tt\footnotesize debasmita.ghose@yale.edu}}%
\thanks{Digital Object Identifier (DOI): see top of this page.}
}
\usepackage{graphicx}
\usepackage{amsmath}
\usepackage{enumitem}
\usepackage{xcolor}
\begin{document}

\maketitle

\begin{abstract}

For effective human-robot collaboration, a robot must align its actions with human goals, even as they change mid-task. Prior approaches often assume fixed goals, reducing goal prediction to a one-time inference. However, in real-world scenarios, humans frequently shift goals, making it challenging for robots to adapt without explicit communication.
We propose a method for detecting goal changes by tracking multiple candidate action sequences and verifying their plausibility against a policy bank. Upon detecting a change, the robot refines its belief in relevant past actions and constructs Receding Horizon Planning (RHP) trees to actively select actions that assist the human while encouraging Differentiating Actions to reveal their updated goal.
We evaluate our approach in a collaborative cooking environment with up to 30 unique recipes and compare it to three comparable human goal prediction algorithms.
Our method outperforms all baselines, quickly converging to the correct goal after a switch, reducing task completion time and improving collaboration efficiency.

\end{abstract}

\begin{IEEEkeywords}
Intention Recognition; Human-Robot Teaming
\end{IEEEkeywords}

\section{Introduction}

\label{sec: intro}
\IEEEPARstart{I}{n} real-world scenarios, humans often change their goals in response to evolving circumstances, new information, or spontaneous decisions. Previous work often addresses changing human goals by relying on explicit communication \cite{pupa2021dynamic,ma2024goal,brawer2023interactive}. While effective, relying on communication assumes humans can and will communicate with the robot, which is often impractical due to physical, situational, or cognitive constraints \cite{hulle2024eyes, ghose2023tailoring,liu2018goal,ghose2025adapting, adamson2021we}. Moreover, in complex tasks, humans may struggle to articulate their goals clearly, leading to misinterpretation. On a busy construction site, for example, a mason and a robot may start with a brick-laying task, the robot assisting by mixing cement and bringing sand, water, and buckets. When a sudden rainstorm halts exterior work, the crew pivots indoors to install drywall panels instead. The materials the robot has already delivered remain useful for rinsing tools and cleaning surfaces, but mixing additional mortar and staging bricks is now counter-productive. Because the noise of heavy machinery and hearing protection prevents verbal updates, the mason cannot notify the robot; only by observing that the worker now reaches for drywall screws rather than bricks can the robot infer the new objective and adapt.
Such goal changes are challenging for robots, especially when there is an overlap in the sequence of actions required to achieve different goals. Without explicitly modeling these goal changes, the robot cannot understand the updated goal effectively or determine which past actions should be retained as relevant context.

\begin{figure}
    \centering
\includegraphics[width=0.8\linewidth]{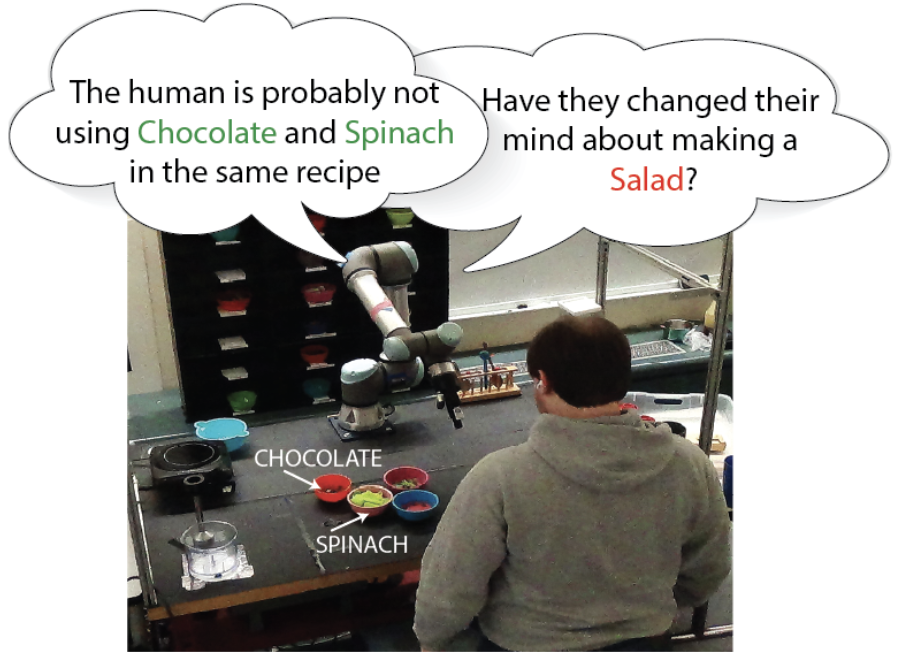}
    \caption{Robot reasoning if the person has changed their mind about making the salad, after observing that the person has picked chocolate.}
    \vspace{-1.5em}
    \label{fig:teaser}
\end{figure}

To that end, we propose a method to explicitly model goal changes without relying on explicit communication. Our approach identifies when likely goal changes occur by tracking multiple candidate action sequences from the history of the collaboration and verifying their plausibility against a policy bank. Once a goal change is detected, the robot refines the set of plausible past actions and constructs Receding Horizon Planning (RHP) trees for each sequence, performing actions that assist the human while encouraging them to take \textit{Differentiating Actions}, which are actions that reveal crucial information about their updated goal.

We evaluate our approach in a collaborative cooking environment (as shown in Fig. \ref{fig:teaser}), both in simulation and on a physical robot. Cooking provides a structured yet realistic setting where goal changes are easily observable.
A human and a robot can cook up to 30 unique recipes in our setup. Our primary point of comparison is the \textit{Recursive Bayesian} \cite{recursive_bayesian} approach that passively models likely goal changes based on observation of human actions. We dynamically manage the relevance of past actions while taking an active approach to human goal prediction, which allows us to outperform the Recursive Bayesian approach across three plausible types of human behavior. Additionally, to illustrate the importance of explicitly modeling goal switches and highlight the magnitude of performance improvement that considering goal switches enables, we benchmark these approaches against two active goal prediction techniques designed for cases where human goals are static: 1) \textit{Critical Decision Points (CDP)}\cite{ghose2024planning}, where a robot actively selects actions to support human goals while influencing them to reveal critical information about their goal, and 2) \textit{Information Gain Maximization }\cite{sadigh2016information}, where a robot actively maximizes its information gain about a human's goal at every state without necessarily helping them achieve it. 

\section{Related Work}





\label{sec: related_work}
In human-robot collaboration, a significant body of work focuses on robots inferring human goals by observing human actions and inferring their intent (for a survey, refer to  \cite{hoffman2024inferring}). Most of this work assumes that human intent remains consistent throughout the collaboration \cite{nikolaidis2018planning,ghose2023tailoring,chang2018effects,hoffman2024inferring,ghose2024planning,laplaza2022context,haninger2022model,pandya2024robots,ghose2025adapting,7759329,belsare2025toward}. 
This assumption does not hold in the real world, as humans can change their minds spontaneously and frequently in response to evolving circumstances and new information. 
Recent methods for handling changing human goals fall into two categories:
\noindent 
\subsubsection{Passive Goal Inference}
Passive approaches rely solely on observing human actions to infer goals \cite{amado2024survey}. Jain and Argall \cite{recursive_bayesian} use recursive Bayesian filters to update goal beliefs, while Murata et al. \cite{murata2019} use gradient descent for dynamic goal adjustments. Both rely on short action histories, making them unsuitable for tasks with overlapping actions. For example, chopping vegetables applies to both a salad and a stew, and these methods lack mechanisms to determine which past actions remain relevant after a goal change. Hiramatsu et al. \cite{10191733} improve on these by explicitly modeling goal transitions. 

\noindent 
\subsubsection{Active Human Influence}
Active approaches aim to influence humans to modify their goals, particularly under uncertainty. Pandya et al. \cite{pandya2024robots} use reach-avoid dynamic games to guide humans toward safer goals, while Pandya et al. \cite{pandya2024towards} extend this line of work by influencing humans to adopt more precise goals. These methods assume humans are uncertain and receptive to influence \cite{sripathy2021dynamically}, which is impractical when humans are confident in their objectives. Moreover, they neglect the action history before the goal change and do not explicitly model goal changes, limiting their utility in tasks with overlapping actions or abrupt changes.

Existing methods for goal inference are limited in handling goal changes. Passive approaches struggle with abrupt changes or overlapping tasks because they rely on incremental updates and short-term action histories. Active approaches fail to address scenarios where humans are certain about their goals and cannot dynamically manage action relevance, either retaining all past actions, causing confusion, or discarding history entirely, losing valuable context.

To address these limitations, we explicitly model goal changes by identifying their precise timing and dynamically managing the relevance of past actions. 
We introduce \textit{Differentiating Actions}, key actions that clarify human intent after a goal change. 
Using Receding Horizon Planning (RHP) \cite{ma2006receding}, we guide humans to reveal updated goals without relying on explicit communication, enabling robust adaptation to complex, real-world collaborative tasks. To validate our approach in a collaborative cooking task, we benchmarked it against several goal-prediction methods.

\section{Method}
\label{sec: method}


We present an algorithm that lets a robot both predict and assist a human partner whose intentions may change mid-task. Consider urban way-finding: a guide robot escorts a pedestrian whose destination is unknown. At each major intersection, the pedestrian’s selected street eliminates most remaining destinations; we call such pivotal turns \textit{Differentiating Actions}. The robot plans its trajectory to influence the human to these informative intersections while staying on routes that best serve the currently most probable destination. Suppose the pedestrian deviates, for example, by turning toward an unexpected café. In that case, the robot interprets this as a goal switch, updates its belief about possible destinations, and adjusts its assistance accordingly.

\subsection{Preliminaries}


The environment's state is represented as $s$, with a shared action space $\mathcal{A}$. The state and the action spaces are fully observable to both the human and robot at all times. The action space contains high-level actions (e.g., in a cooking task, the action space can include boiling water, chopping vegetables, etc.). Both agents have access to a shared action history $\mathcal{H}^{0:t-1}$ up to (but not including) the current time step,
and a complete goal bank $\mathcal{G}$, listing all possible human task goals. We also keep track of the individual action histories of the human and the robot, $\mathcal{H}^{0:t-1}_h$ and $\mathcal{H}^{0:t-1}_r$. Additionally, the robot has a policy bank $\mathcal{P}$, which contains many (but not all) potential policies to achieve the goals in  $\mathcal{G}$. Each policy ($\pi \in \mathcal P$) maps states to actions to progress toward completing a task (e.g., a recipe within a cooking task) for a given goal. We can derive a bi-directional mapping from the policy bank specifying which actions help accomplish which goals.
We assume actions cannot be repeated in an interaction, and collaboration occurs in a turn-taking manner.

The robot aims to reduce uncertainty about a human's current goal and perform actions that support it. At each timestep, the robot infers $g_{pred}$, a probability distribution over the human's possible goals, and at alternate timesteps, selects a robot action $a_r$. 

\subsection{Active Selection of Robot Actions to Reveal Human Goals}
\label{sec: action_selection}

 Like prior work on active goal inference \cite{ghose2024planning,sadigh2016information}, our robot acts to reduce its uncertainty about the human’s goal. It does so by selecting actions that steer the partner to perform \textit{Differentiating Actions} -- human actions that are informative because they advance only a small subset of feasible goals. Actions that aid nearly every remaining goal, or none of them, reveal little; actions that help just a few narrow the robot’s belief and accelerate correct identification of the human's goals early in the interaction.

Similar to Ghose et al. \cite{ghose2024planning}, we formulate the problem of the robot influencing the human to take Differentiating Actions as a discrete-time planning problem, which we solve using Receding-Horizon Planning (RHP) \cite{ma2006receding}. RHP draws from Model Predictive Control \cite{garcia1989model} but focuses on discrete decision-making instead of continuous control. In RHP, an agent solves a planning problem iteratively across a receding horizon, evaluating potential futures by expanding a tree. After planning at a given step, the agent executes the best action, shifts the horizon, and replans for future steps.

\noindent
\subsubsection{Expanding the RHP Tree}
The robot expands an RHP tree to reason about how its future actions could influence the human's subsequent actions, considering the past actions taken by the human. The RHP tree is rooted in the history of past actions taken by the human $\mathcal{H}^{0:t-1}_h$. Each branch of the tree represents a sequence of potential actions that both agents can take from the root node in a turn-taking manner.

\noindent
\subsubsection{Choosing Robot Actions}
\label{sec: choosing_robot_actions}
The robot selects differentiating actions using attractor fields -- a concept initially developed for motion planning \cite{khatib1986real}. The core idea is that certain states (e.g., areas in a map for motion planning) exert attractive or repulsive forces that influence an agent’s decision. Attractive forces pull the agent toward desirable states (e.g., navigating to a goal), while repulsive forces push it away from undesirable states (e.g., avoiding obstacles). 

Based on Jain and Argall \cite{recursive_bayesian}, we adapt this concept from physical space to action space, where certain actions exert an attractive or repulsive force depending on whether they align with a human's goals. We model each goal as such an attractor field over the action space. Intuitively, an attractor field over actions represents which actions are relevant for its corresponding goal. Mathematically, in an action space with $n$ possible actions, a uniform attractor field for a given goal $g$ is represented as $\{0,1\}^n$, where position $i$ is 1 if action $i$ supports $g$ and 0 otherwise. If attractor fields are summed element-wise over a given set of goals, an action will be more attractive if it is relevant to more goals and less attractive if it helps accomplish fewer goals. Crucially, attractor field values are strictly non-negative in our formulation of this concept.

In the context of attractor fields, a differentiating action possesses \textit{low, non-zero attraction}. Actions with zero attraction are irrelevant to any goals under consideration and should be avoided by both the robot and the human. Conversely, if an action is very attractive, it pertains to numerous goals, making it hard for the robot to identify the human's specific goal if such an action is taken by the human.

We define the plausible goals for the history of human actions $\mathcal{H}^{0:t-1}_h$ as the set of goals for which every action in $\mathcal{H}^{0:t-1}_h$ helps accomplish every goal. Intuitively, by aggregating attractor fields across all plausible goals, we effectively create a measure of how helpful each action will be on average, given the action history. This allows the robot to identify and perform actions that are supportive for many plausible goals. As actions cannot be repeated, the robot taking many broadly helpful actions effectively steers the human to take actions that are supportive of fewer goals, which reveals more information about their actual goal.

 For $\mathcal{H}^{0:t-1}_h$, we compute the summed attractor field over its plausible goals similar to \cite{recursive_bayesian}, referred to as $attractor\_field$. 
The initial cost at the root node ($d=0$) of a branch $b$ for an RHP tree is then calculated as:
\begin{align}
\vspace{-1em}
\label{eq: cost_0}
    branch\_costs[b, 0] = -1 \cdot attractor\_field[b[0]],
\end{align}

\noindent
where $b[0]$ is the first action on branch $b$. As the tree expands, the cost of each branch is progressively updated at increasing depths $d$ by subtracting the $attractor\_field$ value for $b[d]$, the action at depth $d$ using: 
\begin{align}
\label{eq: cost}
    branch\_costs[b,d] = branch\_costs[b,d-1] - 
    \\attractor\_field[b[d]]  \notag
\end{align}

Notably, the cost is calculated only at timesteps when the robot would act. If the cost were updated on human timesteps, it would steer the robot towards influencing the human to take high-attraction actions, which would not be differentiating. Conversely, including robot timesteps influences high-attraction actions that are likely to be relevant (satisfying the action-to-goal mapping from $\mathcal{P}$), regardless of the human's goal. Also, we only update the cost for a branch if it was one of the lowest-cost branches at the previous depth, to prevent the robot from performing an irrelevant action earlier in favor of a potentially relevant future action.

\subsection{Identifying Likely Human Goal Changes}
\label{sec: goal_change}

Assume the human starts by following a policy $\pi_{h1}$, which maps states to actions to achieve the goal $g_{h1}$. At some point in the interaction, the human might decide to switch to a different goal $g_{h2}$, adopting a new policy $\pi_{h2}$, where both $g_{h1}$ and $g_{h2}$ belong to the goal bank $\mathcal{G}$.

\noindent
\subsubsection{Tracking Candidate Human Action Sequences}
Since the human’s goal may change at any time, the robot does not assume that every past action remains relevant to the current goal. Instead, the robot tracks candidate human action sequences, denoted as $\mathcal{C}_h$. These sequences are drawn from the complete history of human actions, represented as: $
\mathcal{H}^{0:t-1}_h = \bigl(a_h^{(0)}, a_h^{(2)}, \dots, a_h^{(t-1)}\bigr)$ where each $a_h^{(i)}$ represents an action taken by the human at timestep $i$. Each candidate sequence $c_h \in \mathcal{C}_h$ is a subset of actions in $\mathcal{H}^{0:t-1}_h$, meaning it consists of only some of the human’s past actions. These sequences represent possible actions that might still be relevant to the human’s current most likely goal.
The candidate sequences are chosen so that all actions can be taken while pursuing the given goal. Because different goals may require different subsets of actions, and the robot is uncertain about the human's true goal, the robot needs to track multiple candidate sequences simultaneously. Each sequence reflects a different hypothesis about which actions still matter.
Whenever the human takes a new action $a_h^{(t)}$, the robot updates all candidate sequences by adding this action to each of them. 

\noindent
\subsubsection{Detecting a Goal Change}
To determine whether the human has changed their goal, the robot checks whether the observed human actions still align with any goal reachable by following a policy in the policy bank $\mathcal{P}$. For each candidate action sequence $c_h \in \mathcal{C}_h$, the robot verifies whether it is plausible that a human following a given policy from $\mathcal{P}$ could have taken those actions. This means checking if there exists a policy in $\mathcal{P}$ that could have generated the observed sequence or any of the permutations of actions in the observed sequence for that given goal. 
If at least one candidate sequence matches a policy in $\mathcal{P}$, the robot assumes the human is still following the given goal from the previous timestep.
If all sequences become implausible, it suggests the human's actions may no longer align with any goals in the goal bank $\mathcal{G}$, indicating a likely change to a different goal. In other words, if no tracked subset of human actions ($\mathcal{C}_h$) is consistent with any goal, the robot infers a goal switch; if at least one subset remains consistent with one goal, it cannot be sure a switch occurred. {\color{black} Here, we define a sequence as \textit{consistent} if there is a policy $\pi \in \mathcal{P}$ such that the observed sequence is a subsequence of $\pi$.}



\subsection{Robot Aligning Actions with Updated Goals}
\label{sec: rhp_many}

Once a goal change is detected (as described in Sec. \ref{sec: goal_change}, Fig. \ref{fig:rhp}, Step 1), the robot must reassess which past actions are still relevant and determine the best way to assist the human while reducing uncertainty about the new goal.

\noindent
\subsubsection{Filtering Relevant Human Actions}
The robot aims to identify past actions relevant to the human’s likely new goal. A goal change may render some previous actions useless, prompting the robot to clear its list of relevant robot actions ($\mathcal{H}^{0:t-1}_r$) and find subsets of human actions likely to contribute to the new goal. However, simply considering all possible subsets of past human actions could be computationally intractable, as the number of possible sequences grows exponentially with the number of actions taken. 
The robot ranks actions $a_h^{(i)} \in \mathcal{H}^{0:t-1}_h$ based on \textit{Generality}.
Generality measures how commonly an action appears across different goals in the policy bank $\mathcal{P}$. More general actions (i.e., those that occur in many policies) are more likely to be relevant, even after a goal shift. We progressively eliminate the least general action in an action sequence from $\mathcal{H}^{0:t-1}_h$. This effectively leads the robot to perform a 
{\color{black} ranking and enumeration} over all possible subsets of human actions, prioritizing longer subsets composed of more general actions.
When the robot encounters a subset of actions that could have all been taken while following the same goal, the robot selects this subset, appends the latest human action to it (Fig. \ref{fig:rhp}, Steps 2 and 3), and adds it to $\mathcal{C}_h$. The exploration continues until $\mathcal{C}_h$ reaches length $j$ (or till there are no more subsets to explore). $j$ is a hyperparameter specifying how many candidate sequences to track.


\noindent
\subsubsection{Expanding RHP Trees for Each Plausible Action Sequence} 
\begin{figure}[t!]
    \centering
    \includegraphics[width=0.85\linewidth]{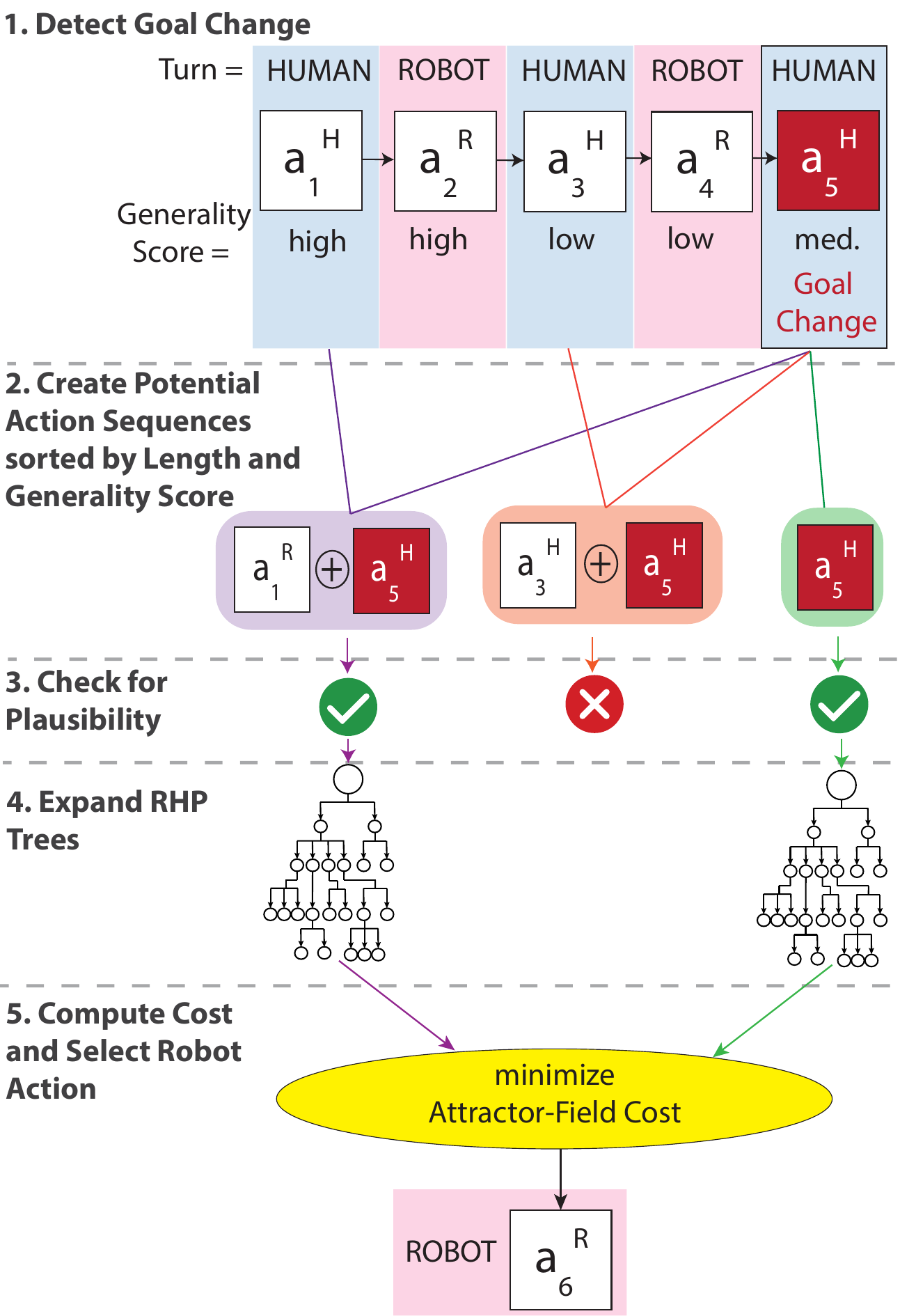}
    \caption{
    \textbf{Steps to Select Robot Action after Goal Change Detection }1. Detect Goal Change if the actions taken during the collaboration don't match any sequences in $\mathcal{P}$. 2. Create potential action sequences by concatenating the action at which the goal change was detected with past actions sorted by increasing generality scores. 3. Check if the potential action sequences are plausible. 4. For all plausible action sequences, expand the RHP trees. 5. Compute attractor field cost and select an optimal next robot action. 
    }
    \vspace{-1.5em}
    \label{fig:rhp}
\end{figure}

Since there can be multiple plausible action sequences ($\mathcal{C}_h$), the robot cannot assume any one sequence is correct since it is uncertain about the human's goal. Instead, it considers multiple plausible interaction histories  ($c_h \in \mathcal{C}_h$) and evaluates potential robot actions under each hypothesis. To do this, the robot constructs an RHP tree for each sequence $c_h$ (following the procedure described in Sec. \ref{sec: action_selection}). Each tree begins at a root node representing one of the plausible candidate sequences $c_h$ and expands by simulating future actions for the human and the robot (Fig. \ref{fig:rhp}, Step 4). 

To decide which action to take that minimizes uncertainty over a human's goals, the robot computes a cost for each branch in every tree, using the attractor field-based cost function from Eqs.\ref{eq: cost_0} and \ref{eq: cost} (Fig. \ref{fig:rhp}, Step 5). As explained in Sec. \ref{sec: choosing_robot_actions}, this cost function aims to steer the human to take differentiating actions early in the interaction. 
The cost update rule from Eq. \ref{eq: cost} is applied only if the branch $b$ had the minimum cost at the previous depth $d-1$. This prevents the robot from prioritizing branches with suboptimal early actions in favor of better actions later, as the belief in the human’s goal may shift before reaching those later actions.
Finally, once the RHP tree is expanded for all plausible action sequences $\mathcal{C}_h$, the branch with the minimum cost summed across all RHP trees is selected. If the same action sequence occurs in multiple trees, we sum their costs, because they are more likely to be relevant regardless of which candidate action sequence supports the true human goal. The robot then takes the first action in that branch and reconstructs the trees after each human response.

\section{Evaluation}
\label{sec: evaluation}

Collaborative cooking scenarios are a standard benchmark for human-robot collaboration algorithms \cite{brawer2023interactive,carroll2019utility,van2022correct,goubard2023cooking}. These tasks often involve overlapping goals, for example, preparing a smoothie or a parfait shares actions like chopping fruits or fetching yogurt, and mirror real-world situations where humans frequently change goals mid-task \cite{brawer2023interactive}. This makes cooking scenarios ideal for evaluating our approach via simulation and physical setups. For such evaluation, we use 30 unique recipes, each of which can serve as the initial goal.
After the switch, the human could adopt any of the remaining 29 recipes as the new goal (we disallow ``switching” to the same recipe). Thus, for every first-goal choice there are 29 distinct second-goal options, giving $30 \times 29 = 870$ pairs of (first goal, second goal).

\subsection{Task and Problem Representation}

We assume that both the human and the robot can perform any step necessary to prepare a recipe and each step takes the same amount of time, regardless of the task or who performs it. We also assume a deterministic environment with a fully observable state space. 
Our experimental setup features a robot and a user in a simplified kitchen environment (Fig. \ref{fig:teaser}). The kitchen includes a blender, a sink with water, a pan, a stove, a serving bowl, a glass, a serving spoon, an eating spoon, a measuring cup, and a storage shelf containing the necessary ingredients. Ingredients are considered non-depletable, meaning a single ingredient can be used in multiple containers. Meals in the experiment could be one of six types: Pasta, Stew, Salad, Oatmeal, Smoothie, or Parfait.



The robot's state and action space are defined using the Planning Domain Definition Language (PDDL) \cite{aeronautiques1998pddl}. We used the PDDLgym library \cite{silver2020pddlgym} to update the environment's state based on actions taken by both the human and the robot. Both the human and the robot could perform the following actions:
\begin{math}
    A = \{\texttt{get(i)}, \texttt{pour(i,d)},  
    \texttt{mix(i)}, \texttt{cook(i)},  \texttt{blend(i)},\\
    \texttt{turn\_on(a)},
    \texttt{collect\_water(d)}, 
       \texttt{reduce\_heat(a)}, \\
      \texttt{serve()}\}
\end{math}
where $i$ is an ingredient, $d$ is a dish such as a bowl, glass, or pan, and $a$ is an appliance. The state space is represented as a set of PDDL literals, with each literal denoting a specific environment feature, like an ingredient’s location (workspace or storage), the state of a container or appliance (e.g., whether the stove is on or off), or the status of a meal (e.g., blended or cooked). {\color{black} Such symbolic abstractions are standard in planning and reasoning \cite{brawer2023fusing}, as they allow tractable inference over long-horizon tasks. In our cooking experiments, the human verbalizes each action, which we map to the appropriate discrete symbol in the policy bank using fuzzy matching. Other non-intrusive mappings are equally feasible, such as visually detecting state changes (e.g., an ingredient placed in the workspace) or using human action recognition models.}


\subsection{Experiments}
In our experiments, the goal for both the human and the robot is to work collaboratively on a combination of two recipes. By combination, we refer to a setup in which the human could start making one recipe and, partway through the interaction, change to another recipe that they complete. In our experiments, only humans know their current goal, and the robot must infer their goal by observing their actions.


We model our simulated human using Clique/Chain Hierarchical Task Networks (CC-HTNs) \cite{hayes2016autonomously}, which can represent both actions that must be performed in sequence and those that can be executed in any order. This makes CC-HTNs ideal for modeling recipes. Each recipe is represented by its own CC-HTN, which the simulated human uses to select actions. We assume that the human always takes the first action, alternating with the robot thereafter.
We evaluate our method using three human models:

\begin{enumerate}[wide=0pt]
\item \textbf{Human with Fixed Goal:} The human follows a single, unchanging goal throughout the interaction (common assumption on human behavior used in literature \cite{ghose2024planning,nikolaidis2017human}). 
\item \textbf{Optimal Human with Goal Changes:} The human may change their goal randomly at any point during the interaction. All actions taken by the human are optimal and correctly aligned with their current goal (a common paradigm used in literature \cite{recursive_bayesian,pandya2024robots}) 
\item \textbf{Suboptimal Human with Goal Changes:} The human may also change their goal randomly during the interaction. However, the human’s actions may include mistakes, where the probability of making mistakes at any step is $p=0.1$. This mode of behavior is studied in \cite{caterino2023human,hopko2022human,kwon2020humans}.
\end{enumerate}

To manage computational constraints and conduct controlled experiments, we ensure human models 2 and 3 experience only one goal change after a certain number of steps by fixing the goal post-change. {\color{black}This restriction to a single switch is not a requirement of our method, but an experimental control to ensure comparability across trials.}

\begin{figure*}
    \centering
    \includegraphics[width=0.85\linewidth]{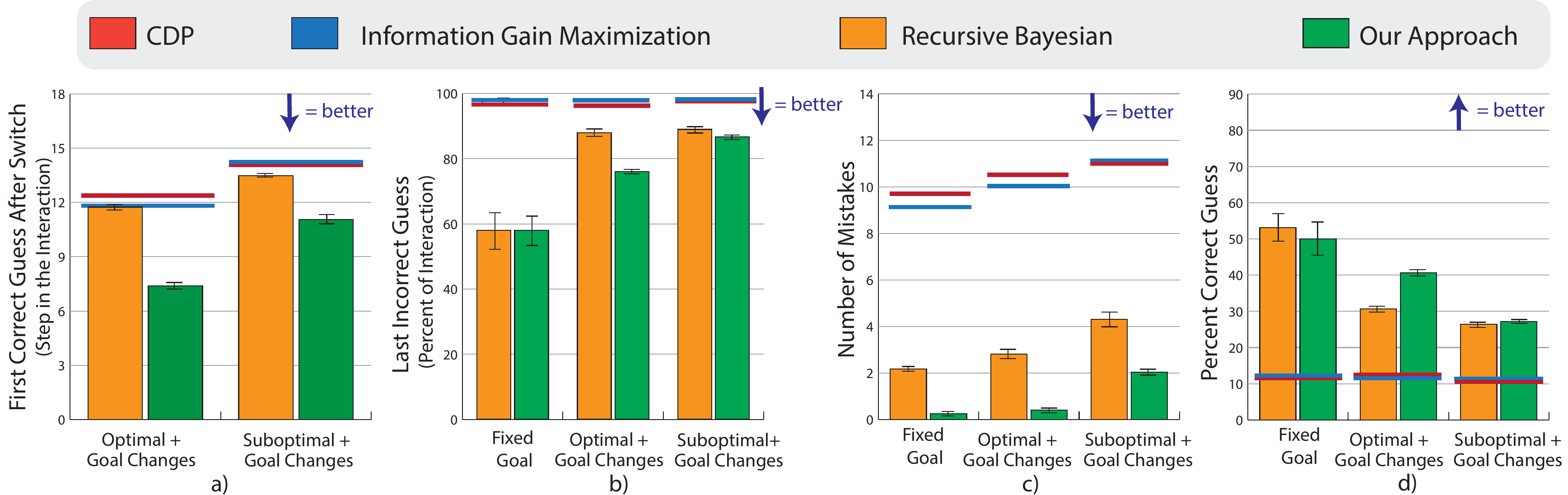}
    \caption{Comparison of performance measures for the proposed method, CDP \cite{ghose2024planning}, Information Gain Maximization \cite{sadigh2016information}, and Recursive Bayesian \cite{recursive_bayesian} in a collaborative cooking simulation. Metrics include (a) first correct guess after a goal switch, (b) last incorrect guess, (c) number of mistakes, and (d) percentage of correct guesses. Results are shown for three simulated human types: stubborn (no goal change), optimal (stochastic goal updates), and suboptimal (stochastic updates with mistakes). In the stubborn condition, all four methods are compared; in the other two conditions, our method is directly compared with Recursive Bayesian, while the red and blue lines indicate the empirical performance bounds provided by CDP and Information Gain Maximization, respectively.}
    \label{fig:results}
    \vspace{-1.5em}
\end{figure*}





\subsection{Implementation Details}
\label{sec: simulation_setup}


\subsubsection{Approximating the Belief Distribution}
Our policy bank $\mathcal{P}$ includes various permutations of action sequences for preparing each goal meal in the goal bank $\mathcal{G}$. Since many recipe steps can be performed in different orders, the number of possible sequences is prohibitively large, making exact computation of the goal distribution infeasible. Instead, we approximate this distribution with a machine learning model. Specifically, we trained a Random Forest classifier on 300,000 sampled sequences (10,000 per recipe across 30 recipes), where each sequence was encoded as a binary presence vector over the action space. The classifier used 200 trees and standard regularization (minimum two samples per leaf, minimum ten samples to split an internal node), yielding 86.8\% accuracy on a held-out 20\% test set. {\color{black}We also experimented with a transformer-based sequence model, which performed comparably but incurred significantly higher latency. Given the real-time demands of our setting, we selected the Random Forest for tractability. Our approach, however, is agnostic to the choice of classifier.}

\noindent
\subsubsection{Pruning the RHP Tree}Due to the large number of valid actions at each state, expanding our full RHP tree to arbitrary depth is computationally prohibitive. We prune the tree by computing the probability of each action based on its preceding sequence using an offline 2-gram approach, which counts how often an action follows a given past action. We then restrict expansion to the $b=5$ most likely actions (branching factor), allowing the tree to reach a depth of $d=3$. To construct candidate human action sequences (Sec. \ref{sec: rhp_many}), we set $j=3$, which specifies the number of candidate sequences tracked. When an action sequence is absent from the n-gram dataset due to the combinatorial growth of unique sequences, we rely on the belief distribution from our random forest model to select branches leading to the most probable goals. {\color{black}In the worst case, expansion of the RHP tree grows as $O(j \times b^d)$. With these pruning parameters ($b=5, d=3, j=3$), runtimes remain tractable, requiring 2–5 seconds per robot action on an NVIDIA RTX 4090.}

\subsection{Comparisons}
To benchmark our approach, we implemented three algorithms in simulation by adapting their original implementations to the collaborative cooking task. Our primary point of comparison is Recursive Bayesian \cite{recursive_bayesian}, designed to handle changing human goals. Additionally, we benchmarked two active goal detection algorithms, Critical Decision Points \cite{ghose2024planning} and Information Gain Maximization \cite{sadigh2016information}, known to perform well for static goals but not for dynamic ones. This comparison establishes empirical performance bounds for our method and Recursive Bayesian, highlighting the value of designing algorithms that explicitly model goal changes.

\noindent
\subsubsection{Recursive Bayesian}
The Recursive Bayesian algorithm \cite{recursive_bayesian} employs Bayesian filtering and a Hidden Markov Model (HMM) to compute goal probabilities. The method assumes that humans can change their goals stochastically during interactions while still taking optimal actions. For a given goal $g$, it calculates the likelihood of the most recent action under $g$ and multiplies it by the HMM state probability, $\sum_{g_{t-1} \in G} P(g_t|g_{t-1}) \cdot b(g_{t-1})$. For action selection, actions are modeled as attractor-repeller fields; an action $a$ attracts $g$ if it aids in achieving $g$, and repels otherwise. They represent $a$ as a vector with an entry of 1 for an attractor and -1 for a repeller for each goal. Fields are generated from the most recent human action and candidate robot actions, and the robot selects the action most similar to the human action based on cosine similarity.
This differs from our approach, as they construct attractor fields per-action, whereas we construct an attractor field for each goal. Additionally, we directly use attractor field values for action selection, whereas they use them to compute similarity between the potential next robot actions and the most recent human action.

\noindent
\subsubsection{Critical Decision Points}
In Critical Decision Points (CDP) \cite{ghose2024planning}, the robot identifies states where observing human actions yields insight into their goals, treating each state independently and ignoring temporal dependencies. It evaluates the information gained in each state and selects actions to influence humans to reach that state by expanding an RHP tree. The robot observes the human's response and updates its belief about its goal by executing the first action that minimizes the cost. It then lets the human respond with a valid HTN action, expands the tree from the new state, and iteratively updates its goal probabilities.



\noindent
\subsubsection{Information Gain Maximization}
The Information Gain Maximization approach \cite{sadigh2016information} enables a robot to explore a human's internal state by planning actions that maximize expected information gain. We adapt this by expanding an RHP tree to depth 3, where the robot optimizes a cost function with two terms: one maximizing the magnitude of entropy loss between the root and leaf nodes to reduce ambiguity, and the other rewarding actions aligned with the most likely goal while penalizing deviations. Goal probabilities are computed as detailed in Sec. \ref{sec: simulation_setup} and the RHP tree is expanded for the combined cost as explained for Critical Decision Points.


\noindent







\subsection{Performance Measures}
\label{sec: metrics}
We compare the performance of our method against the baselines for our proposed task using the following measures. 
\begin{enumerate}[wide=0pt]
    \item \textbf{First Correct Guess After Switch: } The number of timesteps taken to correctly identify the human's goal after a goal change (irrelevant to the stubborn human model). 
    \item \textbf{Last Incorrect Guess: } Percentage of the interaction that had elapsed when the robot incorrectly guessed the human's goal for the last time. While this measure considers both goals for human models 2 and 3, the last incorrect guess almost always occurs during the second goal.
    \item \textbf{Number of mistakes: } The total extra steps taken by both agents to finish the interaction, compared to the ground-truth steps from the respective HTN. 
    \item \textbf{Percentage of Correct Guesses: } Percentage of timesteps when the robot correctly identifies the human's goals. 
\end{enumerate}
\section{Results}
\label{sec: results}

\subsubsection{Simulation}

Fig. \ref{fig:results} quantitatively compares our proposed method with three baseline approaches across several key measures and three human models in the simulated collaborative cooking task (as described in Sec. \ref{sec: evaluation}). Mean and standard errors were calculated across 870 recipe combinations and averaged over three trials. For clarity, we don't report standard error bars for the empirical performance bounds. When the human makes no mistakes, the recipe combinations had an average length of 28.7 steps.

Fig. \ref{fig:results} a) demonstrates that our approach accurately identifies the human’s goal relatively quickly after detecting a goal switch, especially when compared to other algorithms across both relevant human models.  However, this metric can be somewhat noisy, as all methods might occasionally predict the correct goal by chance after a goal switch without resolving uncertainty about the goal.

Therefore, we compute the last incorrect guess metric, as it marks the point in the interaction where the robot no longer makes erroneous predictions about the human’s goal. Fig. \ref{fig:results} b) shows that our approach and Recursive Bayesian approach generally achieve this around the midpoint of the sequence when the human does not switch goals earlier than the other two baselines. However, our approach outperforms all baselines when the human switches goals stochastically. This shows that our method can resolve uncertainty about the human's goals faster than other algorithms, leading to a higher percentage of correct guesses, as shown by Fig. \ref{fig:results} d) when the human switches goals stochastically.

Fig. \ref{fig:results} c) shows the total extra steps to complete the collaboration. If the robot performs every action incorrectly, the interaction could take twice as long, as the human would need to perform every correct step towards task completion. Conversely, if the robot correctly assists the human, the task can be completed in 28.7 steps, the average length of a ground truth sequence. The number of extra steps in our approach is minimal across all three human models, showing that our approach can help maximize collaboration efficiency.

\begin{figure}[t]
    \centering
    \includegraphics[width=1\linewidth] {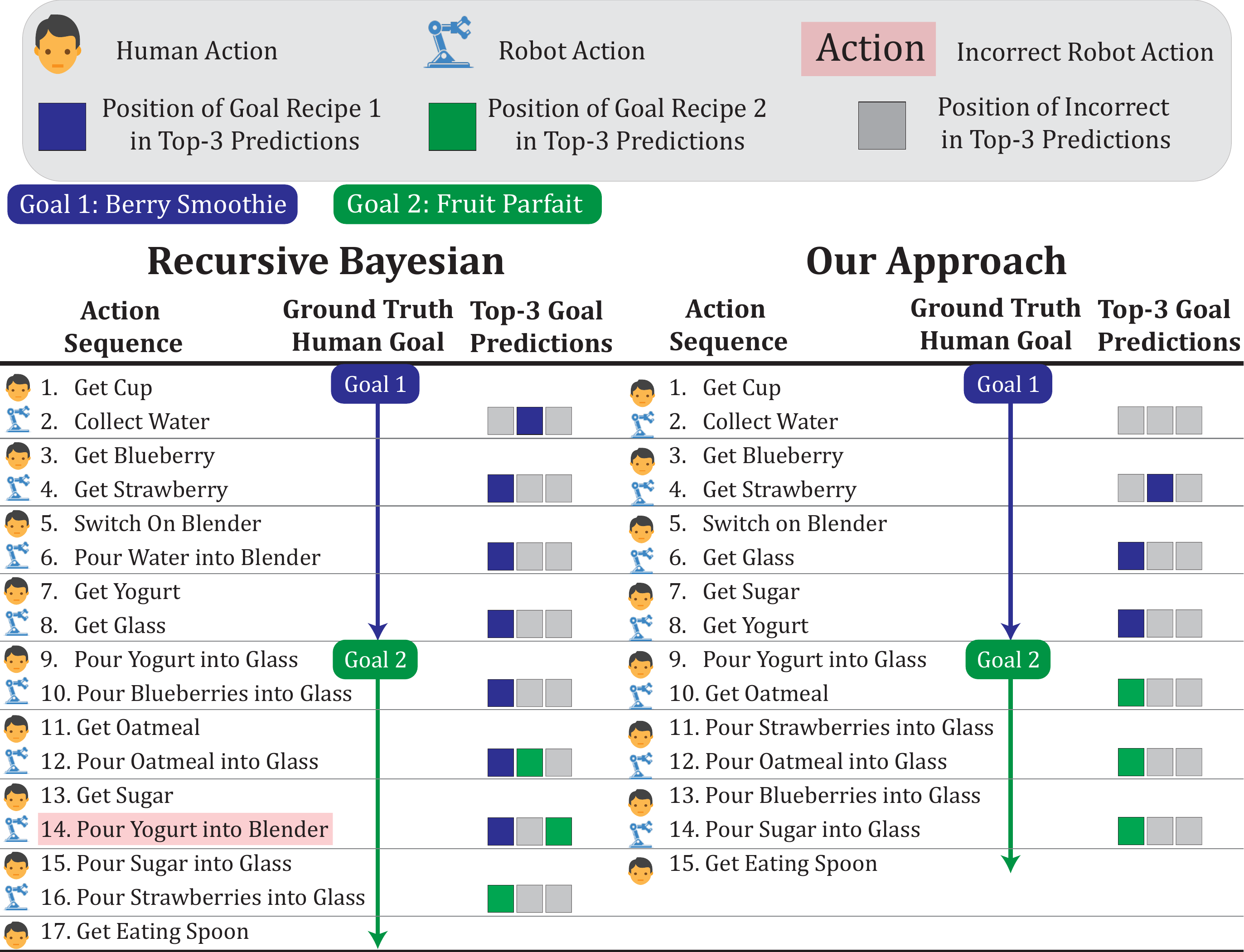}
    \caption{Results of the Case Study with the
    Physical Robot running Recursive Bayesian 
    \cite{recursive_bayesian} and Our Proposed 
    Algorithm: A group 
    of columns represent the method. For each group, 
    the first column depicts the agent taking action, 
    the second column depicts the action sequences 
    taken by the respective method, the third column 
    denotes the ground truth human goal, and the 
    fourth column the presence and position of the 
    current ground truth goal recipe in the top-3 
    most probable recipes from the goal prediction 
    model (left-most being the most probable recipe).}
    \vspace{-1em}
    \label{fig: case_study}
\end{figure}

\subsubsection{Physical Robot}

We conducted a proof-of-concept study using a real robot and a human collaborator (one of the researchers) to compare our method with Recursive Bayesian \cite{recursive_bayesian} in a collaborative cooking task (Fig. \ref{fig: case_study}; see Supplementary Video). For each robot step, we checked whether the correct recipe goal appeared among the top three predictions. To ensure fairness, the human performed the same initial actions in both methods and changed goals at the same timestep by taking an identical revealing action without mistakes. In the study, the human initially planned a Berry Smoothie but switched to a Fruit Parfait. In our approach, after berries were fetched and the blender was switched on, the robot supported the initial goal by fetching yogurt and a glass, then quickly detected the switch when the human added yogurt to the glass instead of the blender, and subsequently assisted by fetching oatmeal and adding fruits. Recursive Bayesian correctly identified the initial goal but struggled to update after the change until the final step.

\section{Discussion}
\label{sec: discussion}

Our approach relaxes many prior assumptions to reflect real-world human-robot collaboration better. A large body of research work assumes fixed human goals 
 \cite{hoffman2024inferring}, optimal human behavior \cite{ghose2025adapting}, and non-overlapping actions between goals \cite{wu2021too}. In contrast, our approach can handle changing human goals, suboptimal actions, and many overlapping actions between different goals. We achieve this by explicitly modeling when a goal shift occurs by tracking multiple different sets of actions that can be plausibly taken towards a given goal. Our approach expands multiple RHP trees to track potential futures for all plausible sequences simultaneously by utilizing our novel attractor-field-based cost function, which models the relevance of future actions related to past human actions. This effectively selects actions that reduce the robot's uncertainty about the human's goals while concurrently supporting them in achieving the most likely outcomes, regardless of whether the human acts optimally. 

Two of our baselines, CDP \cite{ghose2024planning}, and Information Gain Maximization \cite{sadigh2016information}, struggle with dynamic goal changes despite actively influencing the humans to reveal their goals as evident in our results (Fig. \ref{fig:results}). They determine the most likely goal at each timestep based solely on current observations, ignoring how human actions evolve, often missing goal changes. Therefore, we don't use these methods as a direct comparison to our work, but instead to serve as empirical performance bounds for our method.
On the contrary, Recursive Bayesian \cite{recursive_bayesian} captures the progression of actions under changing goals but relies on a short action history. This works well when the human's goals remain unchanged, but it overlooks valuable context that could help the robot track past relevant human actions. Moreover, this approach relies on passive observation of human behavior instead of actively influencing human behavior to help the robot reduce uncertainty over human goals.{\color{black} While this is not a fundamental limitation of Bayesian inference, it means that Bayesian approaches must wait for distinguishing actions to occur naturally, which can take significantly longer. Our method differs by enabling the robot to take differentiating actions early, actively disambiguating overlapping goals and thereby accelerating recognition after a change (see Fig. \ref{fig:results}a)).}

We acknowledge several limiting assumptions that may not apply in the real world. Our approach requires an explicit goal and policy bank, assumes tasks are represented in a discrete space, and uses a turn-based interaction model where neither agent can ``wait" and actions, once taken, cannot be repeated. Future work will relax these constraints to enable human goal prediction without a predefined goal bank.




\bibliographystyle{IEEEtran}
\bibliography{references}

\end{document}